\documentclass[runningheads]{llncs}
\usepackage[T1]{fontenc}
\usepackage{graphicx}
\usepackage{booktabs}
\usepackage[misc]{ifsym}
\newcommand{\corr}{(\Letter)}
\usepackage{amsmath,amssymb,amsfonts}
\usepackage{multirow}
\usepackage{cite}
\usepackage[hidelinks]{hyperref}

\begin{document}

\title{HyperLabel: Multi-Label Classification via Hypergraph-Based Label Correlation Modeling}

\titlerunning{HyperLabel: Hypergraph-Based Multi-Label Classification}

\author{Peiyu Zhang\inst{1} \and
Heng Ping\inst{1} \and
Nikos Kanakaris\inst{2} \and
Yucheng Zhao\inst{3} \and
Shixuan Li\inst{1} \and
Wei Yang\inst{1} \and
Xiongye Xiao\inst{3} \and
Paul Bogdan\inst{1} \corr}
\authorrunning{P. Zhang et al.}

\institute{University of Southern California\\
\email{\{pzhang65,hping,sli97750,wyang930,pbogdan\}@usc.edu}
\and
Amazon, AWS\\
\email{nikosk@amazon.com}
\and
University of Tennessee, Knoxville\\
\email{yzhao106@vols.utk.edu}, \email{xxiao9@utk.edu}}

\maketitle

\begin{abstract}
Multi-label classification (MLC) requires predicting multiple relevant labels for each instance, where a central challenge is modeling complex label dependencies arising from co-occurrence patterns. Existing approaches are limited in capturing high-order label correlations, relying on implicit learning through contrastive objectives or pairwise attention mechanisms without structural guidance. We propose HyperLabel, an encoder-decoder framework that explicitly models label dependencies through hypergraph neural networks. Our contributions are twofold: \textbf{(i)} We construct a label hypergraph where sample-defined hyperedges naturally encode multi-way co-occurrence patterns, providing explicit structural prior knowledge that captures relationships beyond pairwise interactions. \textbf{(ii)} We propose a unified cross-modal learning approach where HGNN+ performs bidirectional message passing to integrate feature information with label structure, and a shared cross-attention decoder processes both modalities through complementary learning objectives. Extensive experiments on seven benchmark datasets demonstrate that HyperLabel achieves state-of-the-art performance, with particularly significant improvements on macro-F1 scores (+10.3\% on Delicious, +8.2\% on Bibtex), validating that explicit hypergraph structure effectively captures complex label relationships. The code is available at \url{https://github.com/iZHpy/Multi-label_hypergraph}.

\end{abstract}

\section{Introduction}
Multi-label classification (MLC) is a fundamental learning problem in which each instance may belong to multiple categories simultaneously. It arises in a wide range of applications, including text categorization, image understanding, bioinformatics, medical diagnosis, and web content tagging~\cite{gibaja2014multi,liu2017deep}. For example, a news article may involve multiple topics such as politics, economy, and international relations, while a product may be associated with several attributes such as fashion, vintage, and sustainable~\cite{prabhu2018extreme,you2019attentionxml,mcauley2015image,he2016ups}. The prevalence of such multi-faceted data makes effective multi-label classification an important and widely studied problem.

Unlike single-label classification, MLC requires predicting multiple labels simultaneously for each instance, introducing the additional challenge of learning intricate dependencies among the labels themselves. Multiple labels assigned to the same sample often exhibit rich co-occurrence patterns. For instance, in wildlife monitoring, ``forest habitat'' and ``nocturnal activity'' tend to appear together for certain species, while ``aquatic environment'' and ``migratory behavior'' form another common pattern. Effectively capturing these complex label correlations is crucial for achieving accurate multi-label predictions.

Existing approaches to address these challenges fall into several categories. Probabilistic classifier chain (PCC) methods model joint label dependencies through sequential prediction with autoregressive architectures~\cite{nam2017maximizing}, but suffer from scalability bottlenecks, error accumulation, and sensitivity to predetermined label ordering. VAE-based methods learn shared probabilistic latent spaces for features and labels~\cite{bai2020disentangled,zhao2021hot,bai2022gaussian}, where MPVAE~\cite{bai2020disentangled} encodes pairwise label relationships via a multivariate probit model, while HOT-VAE~\cite{zhao2021hot} and C-GMVAE~\cite{bai2022gaussian} attempt to capture higher-order patterns through attention-based message passing and contrastive learning, respectively. However, these methods depend on predefined distributional assumptions and rely on implicit learning mechanisms that require extensive hyperparameter tuning. Graph-based methods such as LaMP~\cite{lanchantin2019neural} construct label graphs and apply attention-based message passing to learn dependencies, yet operate without explicit prior knowledge about sample-label correspondences, limiting their capacity to capture multi-way label interactions beyond pairwise relationships.

To address these limitations, we propose \textbf{HyperLabel}, an encoder-decoder framework that models complex label dependencies through hypergraph neural networks~\cite{feng2019hypergraph,gao2022hypergraph}. Our key insight is to leverage the natural correspondence between samples and their label sets to construct a hypergraph where labels serve as nodes and samples define hyperedges. This design encodes sample-label relationships as explicit structural prior knowledge, naturally represents multi-way interactions where a single sample connects multiple labels through one hyperedge~\cite{heo2022hypergraph}, and provides structural inductive bias~\cite{jin2020graph,wu2020comprehensive}. HGNN+ serves as the label encoder, performing bidirectional message passing between label nodes and sample hyperedges to integrate feature information with label representations while capturing multi-way correlations. Within the encoder-decoder architecture, sample features pass through a Transformer encoder to obtain latent representations, while labels are encoded via HGNN+ to produce embeddings enriched with both feature context and structural dependencies. A shared decoder processes both embedding types, enabling cross-modal information exchange through parameter sharing. Multiple complementary losses further strengthen feature-label and label-label correlation learning from diverse perspectives.

Our main contributions are summarized as follows:
\begin{itemize}
\item \textbf{Hypergraph Construction:} A novel hypergraph construction is introduced to leverage sample-label correspondences as prior structural knowledge, enabling effective modeling of both feature-label and label-label relationships through explicit graph topology.
\item \textbf{HGNN+ Label Encoder:} HGNN+ is employed as a label encoder that integrates feature information into label representations while capturing high-order label correlations through bidirectional hypergraph message passing.
\item \textbf{Multi-Objective Learning:} A comprehensive learning framework incorporating alignment, reconstruction, classification, and contrastive losses is designed to model feature-label and label-label interactions from multiple perspectives.
\item \textbf{Comprehensive Evaluation:} Extensive experiments on seven benchmark datasets demonstrate that HyperLabel achieves state-of-the-art performance, validating the effectiveness of the proposed hypergraph-based approach.
\end{itemize}

\section{Related Work}
Multi-label classification has evolved from traditional problem transformation methods to sophisticated deep learning approaches~\cite{tsoumakas2007multi,gibaja2014multi}. The most relevant work can be organized into three categories: sequential prediction methods, latent space learning, and graph-based models, followed by a discussion of hypergraph neural networks.

\textbf{Sequential Prediction Methods.} Probabilistic classifier chains model label dependencies through sequential prediction, where each label is conditioned on previously predicted labels~\cite{nam2017maximizing}. Deep learning extensions employ RNNs and sequence-to-sequence architectures~\cite{wang2016cnn}, but their autoregressive nature prevents parallelization, causes error propagation, and requires predefined label ordering~\cite{gibaja2014multi}.

\textbf{Latent Space Learning.} VAE-based methods learn probabilistic latent representations encoding both features and labels~\cite{kingma2013auto}. MPVAE~\cite{bai2020disentangled} learns aligned Gaussian spaces with multivariate probit decoders for pairwise label correlations. HOT-VAE~\cite{zhao2021hot} and C-GMVAE~\cite{bai2022gaussian} extend this with attention-based message passing and contrastive learning for higher-order dependencies, respectively. Despite their effectiveness, these methods rely on implicit mechanisms requiring extensive hyperparameter tuning to discover multi-way interactions~\cite{khosla2020supervised}.

\textbf{Graph-Based Methods.} Graph neural networks construct label graphs where nodes represent classes and edges encode relationships~\cite{chen2019multi,kipf2017semi,lanchantin2019neural}. LaMP~\cite{lanchantin2019neural} applies attention-based message passing on fully-connected label graphs for adaptive learning of label importance. However, lacking explicit structural priors, these models must learn all relationships from scratch~\cite{zhang2020deep}.

\textbf{Hypergraph Neural Networks.} Hypergraphs generalize graphs by allowing hyperedges to connect arbitrary numbers of nodes, naturally modeling multi-way relationships~\cite{feng2019hypergraph}. Hypergraph neural networks perform bidirectional message passing between nodes and hyperedges~\cite{bai2021hypergraph,gao2022hypergraph}, with applications spanning visual question answering~\cite{heo2022hypergraph}, text classification~\cite{ding2020be}, and image understanding~\cite{bai2021hypergraph}. Despite the strong conceptual alignment between hyperedges and multi-label samples, hypergraph neural networks have not been explored for modeling sample-label correspondences in multi-label classification.

Our work bridges these directions by explicitly encoding sample-label relationships as structural prior knowledge. Unlike graph-based methods lacking sample structure~\cite{lanchantin2019neural,chen2019multi} or VAE-based methods learning correlations implicitly~\cite{bai2022gaussian,zhao2021hot}, our approach models both feature-label and label-label dependencies through unified hypergraph representation, effectively capturing high-order correlations.

\begin{figure}[t] 
    \centering
    \includegraphics[width=\textwidth]{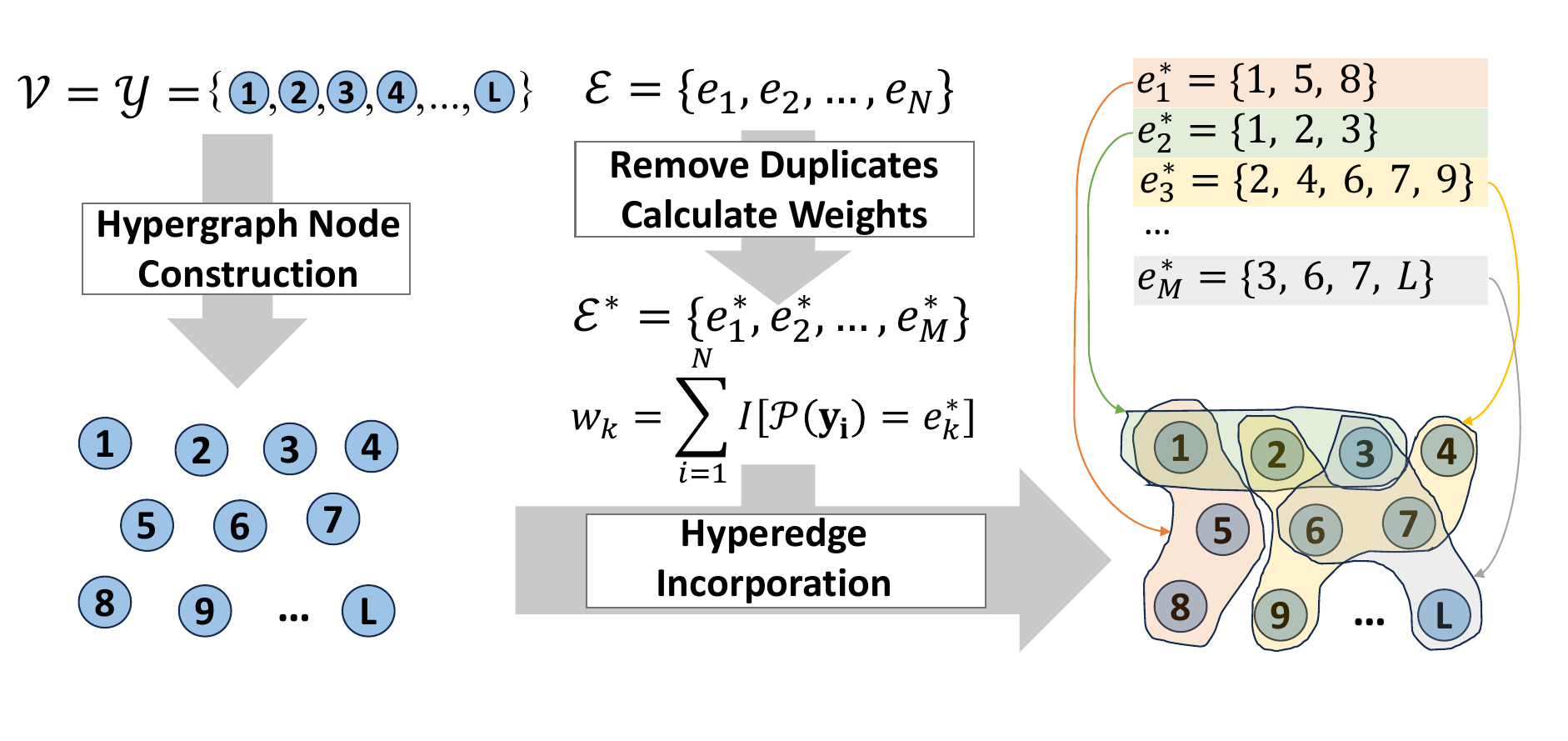} 
    \caption{Hypergraph construction process. Label nodes $\mathcal{V} = \{1, 2, \ldots, L\}$ are connected by sample-defined hyperedges. Each sample's positive label set $\mathcal{P}(\mathbf{y}_i)$ defines a hyperedge (e.g., $e_1^* = \{1, 5, 8\}$). Duplicate hyperedges are merged and assigned weights $w_k$ based on occurrence frequency. The resulting hypergraph explicitly encodes multi-way label co-occurrence patterns.}
    \label{fig:hypergraph}
\end{figure}

\section{Proposed Method}
In this section, we present the HyperLabel framework for multi-label classification. Our approach consists of four key components: (a) \textbf{Hypergraph Construction} encodes sample-label correspondences as explicit structural prior knowledge. (b) \textbf{HyperLabel Encoder} comprises a feature encoder based on Transformers and a label encoder based on hypergraph neural networks. (c) \textbf{HyperLabel Decoder} is a shared network that processes both feature and label embeddings to enable cross-modal information exchange. (d) \textbf{Learning Objectives} integrate multiple complementary losses to strengthen feature-label and label-label correlation learning.

\subsection{Problem Formulation}

In multi-label classification, we are given a dataset $\mathcal{D} = \{(\mathbf{x}_i, \mathbf{y}_i)\}_{i=1}^{N}$ containing $N$ samples. Each input $\mathbf{x}_i \in \mathbb{R}^{D}$ is a $D$-dimensional feature vector, and each output $\mathbf{y}_i \in \{0,1\}^{L}$ is a binary vector representing the presence or absence of $L$ label classes. Let $\mathcal{Y} = \{1, 2, \ldots, L\}$ denote the label index set. The positive label set for sample $i$ is $\mathcal{P}(\mathbf{y}_i) = \{j \in \mathcal{Y} : y_{ij} = 1\}$. The objective is to learn a mapping function $f_{\theta}$:
\begin{equation}
f_{\theta}: \mathbf{x} \mapsto \hat{\mathbf{y}}, \quad \mathbf{x} \in \mathbb{R}^{D}, \hat{\mathbf{y}} \in [0,1]^{L}
\end{equation}
where $\hat{\mathbf{y}} = f_{\theta}(\mathbf{x})$ represents predicted probability scores for each label.

\subsection{Hypergraph Construction}

The core innovation of HyperLabel lies in constructing a hypergraph that explicitly encodes the natural correspondence between samples and their associated label sets, providing structural prior knowledge that guides the learning of both feature-label and label-label correlations.

We define a hypergraph $\mathcal{G} = (\mathcal{V}, \mathcal{E}, \mathbf{W})$, where $\mathcal{V}$ is the set of nodes, $\mathcal{E}$ is the set of hyperedges, and $\mathbf{W} \in \mathbb{R}^{|\mathcal{E}|}$ is a vector of hyperedge weights. Each label class corresponds to a node, thus $\mathcal{V} = \mathcal{Y} = \{1, 2, \ldots, L\}$. Each training sample defines a hyperedge connecting its positive labels:
\begin{equation}
e_i = \mathcal{P}(\mathbf{y}_i) = \{j \in \mathcal{Y} : y_{ij} = 1\}
\end{equation}
This construction naturally captures multi-way relationships, as a hyperedge $e_i$ directly connects all labels that co-occur in sample $i$.

Multiple samples may share identical label sets, resulting in duplicate hyperedges. Let $\mathcal{E}^* = \{e_1^*, e_2^*, \ldots, e_M^*\}$ denote the set of $M$ unique hyperedges after removing duplicates. Each unique hyperedge $e_k^*$ is assigned a weight reflecting its frequency:
\begin{equation}
w_k = \sum_{i=1}^{N} \mathbb{I}[\mathcal{P}(\mathbf{y}_i) = e_k^*]
\end{equation}
where $\mathbb{I}[\cdot]$ is the indicator function. The hypergraph structure is represented by an incidence matrix $\mathbf{H} \in \{0,1\}^{L \times M}$, where $H_{jk} = 1$ if node $j \in e_k^*$ and $H_{jk} = 0$ otherwise. An illustration of this process is shown in Figure~\ref{fig:hypergraph}.

This construction offers several advantages. It explicitly encodes sample-label correspondences as structural prior knowledge, reducing the need to learn these relationships purely from gradients. Hyperedges naturally represent multi-way label interactions beyond pairwise dependencies. Moreover, when many samples share label sets, $M$ can be much smaller than $N$, and the weighted hypergraph provides a compact yet informative representation.

\begin{figure*}[t]
    \centering
    \makebox[\textwidth][c]{\includegraphics[width=1.05\textwidth]{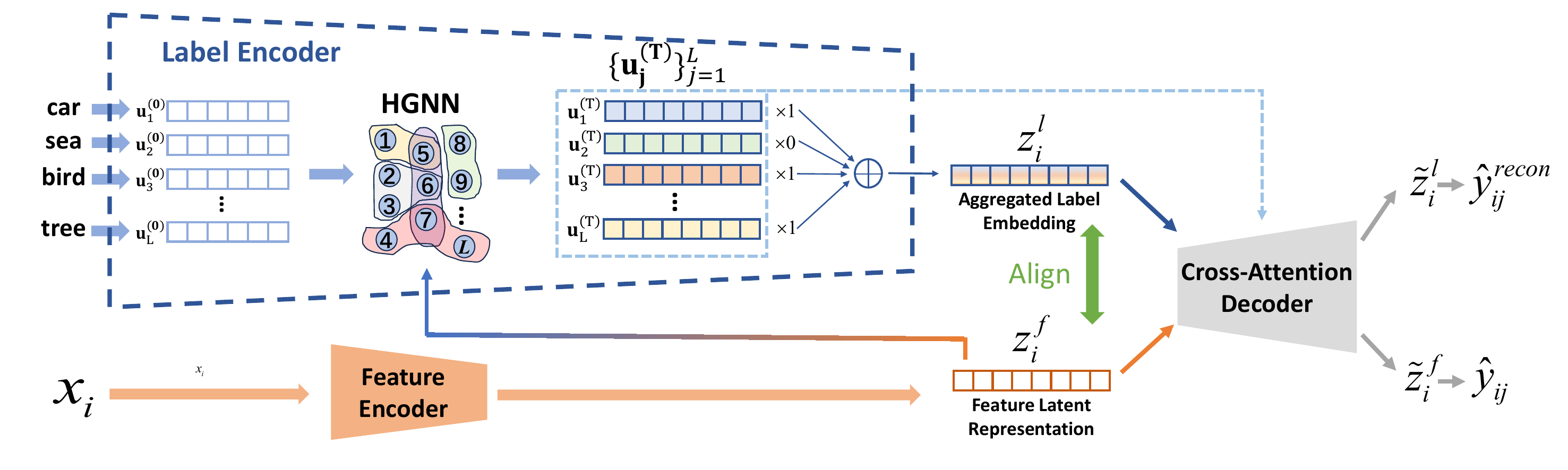}
    }
    \caption{Overview of HyperLabel framework. Labels are encoded via HGNN+ with hypergraph message passing, while features pass through a Transformer encoder. Aggregated label embeddings are aligned with feature embeddings and both are processed by a shared cross-attention decoder to produce label predictions $\hat{y}_{ij}$ and reconstructed predictions $\hat{y}_{ij}^{\text{recon}}$.}
    \label{fig:hyperlabel_framework}
\end{figure*}

\subsection{HyperLabel Encoder}

The HyperLabel encoder comprises two parallel pathways, as illustrated in Figure~\ref{fig:hyperlabel_framework}. The feature encoder employs a Transformer architecture, while the label encoder leverages HGNN+ to integrate structural knowledge from the hypergraph with feature information.

\subsubsection{Feature Encoder}

Given an input sample $\mathbf{x}_i \in \mathbb{R}^{D}$, the feature encoder produces a latent representation through Transformer-based encoding~\cite{vaswani2017attention}:
\begin{equation}
\mathbf{z}_i^f = \text{Transformer}(\mathbf{x}_i; \theta_f) \in \mathbb{R}^{d}
\label{eq:feature_encoder}
\end{equation}
where $\theta_f$ denotes the Transformer parameters and $d$ is the embedding dimension.

\subsubsection{Label Encoder via HGNN+}

The label encoder operates on the constructed hypergraph $\mathcal{G} = (\mathcal{V}, \mathcal{E}^*, \mathbf{W})$ to learn label representations enriched with both structural dependencies and feature context, employing HGNN+~\cite{gao2022hypergraph} for bidirectional message passing between nodes and hyperedges.

\textbf{Initial Label Embeddings.} Each label $j \in \mathcal{Y}$ is initialized with a learnable embedding:
\begin{equation}
\mathbf{u}_j^{(0)} = \text{MLP}_{\text{label}}(j; \theta_l) \in \mathbb{R}^{d}
\label{eq:initial_label}
\end{equation}

\textbf{Hyperedge Embeddings.} For each unique hyperedge $e_k^*$, we aggregate feature embeddings from all samples sharing this label configuration. Let $\mathcal{S}_k = \{i : \mathcal{P}(\mathbf{y}_i) = e_k^*\}$ denote the corresponding sample indices. The hyperedge embedding is computed via attention-weighted aggregation:
\begin{equation}
\mathbf{h}_k^{(0)} = \sum_{i \in \mathcal{S}_k} \alpha_{ki} \mathbf{z}_i^f, \quad \alpha_{ki} = \frac{\exp(\mathbf{q}_k^\top \mathbf{z}_i^f)}{\sum_{j \in \mathcal{S}_k} \exp(\mathbf{q}_k^\top \mathbf{z}_j^f)}
\label{eq:hyperedge_init}
\end{equation}
where $\mathbf{q}_k$ is a learnable query vector for hyperedge $e_k^*$, allowing the model to differentially weight samples' contributions and effectively incorporate feature information into the hypergraph structure.

\textbf{Bidirectional Message Passing.} HGNN+ performs iterative message passing to refine both label and hyperedge representations. At iteration $t$, the process alternates between two propagation directions:

\textit{Node-to-Hyperedge Propagation:} Label nodes send messages to their incident hyperedges:
\begin{equation}
\mathbf{m}_k^{(t)} = \sum_{j \in e_k^*} \frac{1}{\sqrt{d_j \cdot |e_k^*|}} \mathbf{u}_j^{(t)}
\label{eq:node_to_edge}
\end{equation}
where $d_j = \sum_{k=1}^{M} H_{jk}$ is the degree of label node $j$.

\textit{Hyperedge-to-Node Propagation:} Hyperedges propagate aggregated messages back to label nodes with residual connections:
\begin{equation}
\mathbf{u}_j^{(t+1)} = \sigma\left(\mathbf{u}_j^{(t)} + \sum_{k: j \in e_k^*} \frac{w_k}{\sqrt{d_j \cdot |e_k^*|}} \left(\mathbf{W}_h \mathbf{m}_k^{(t)} + \mathbf{W}_e \mathbf{h}_k^{(t)}\right)\right)
\label{eq:edge_to_node}
\end{equation}
where $\sigma$ is a non-linear activation function, $\mathbf{W}_h, \mathbf{W}_e \in \mathbb{R}^{d \times d}$ are learnable transformation matrices, and the hyperedge weight $w_k$ modulates message strength based on label co-occurrence frequency.

After $T$ iterations, we obtain enriched label embeddings $\{\mathbf{u}_j^{(T)}\}_{j=1}^{L}$ that encode both structural dependencies and feature context.

\textbf{Sample-Specific Label Aggregation.} To establish correspondence between feature and label representations at the sample level, we aggregate label embeddings associated with each sample's positive labels:
\begin{equation}
\mathbf{z}_i^l = \sum_{j \in \mathcal{P}(\mathbf{y}_i)} \mathbf{u}_j^{(T)}
\label{eq:label_aggregation}
\end{equation}
This produces a sample-specific label embedding $\mathbf{z}_i^l \in \mathbb{R}^{d}$ that is directly comparable with the feature embedding $\mathbf{z}_i^f$, enabling effective alignment between the latent feature space and latent label space.

\subsection{HyperLabel Decoder}

The decoder processes both feature and label embeddings through a shared cross-attention architecture. It takes the aggregated embeddings ($\mathbf{z}_i^f$ and $\mathbf{z}_i^l$) as queries and attends to the individual label embeddings $\{\mathbf{u}_j^{(T)}\}_{j=1}^{L}$ to produce task-specific outputs.

\subsubsection{Cross-Attention Mechanism}

For an input query embedding $\mathbf{q} \in \mathbb{R}^{d}$ (either $\mathbf{z}_i^f$ or $\mathbf{z}_i^l$), the cross-attention operation is:
\begin{equation}
\text{CrossAttn}(\mathbf{q}; \{\mathbf{u}_j^{(T)}\}_{j=1}^{L}) = \sum_{j=1}^{L} \beta_j (\mathbf{W}_v \mathbf{u}_j^{(T)})
\label{eq:cross_attention}
\end{equation}
where the attention weights are:
\begin{equation}
\beta_j = \frac{\exp((\mathbf{W}_q \mathbf{q})^\top (\mathbf{W}_k \mathbf{u}_j^{(T)}) / \sqrt{d})}{\sum_{k=1}^{L} \exp((\mathbf{W}_q \mathbf{q})^\top (\mathbf{W}_k \mathbf{u}_k^{(T)}) / \sqrt{d})}
\label{eq:attention_weights}
\end{equation}
Here, $\mathbf{W}_q, \mathbf{W}_k, \mathbf{W}_v \in \mathbb{R}^{d \times d}$ are learnable transformation matrices shared across both processing pathways.

\subsubsection{Dual Processing Pathways}

\textbf{Label Reconstruction Pathway.} When processing the aggregated label embedding $\mathbf{z}_i^l$, the decoder reconstructs label predictions through cross-attention followed by inner products with individual label embeddings:
\begin{equation}
\tilde{\mathbf{z}}_i^l = \text{CrossAttn}(\mathbf{z}_i^l; \{\mathbf{u}_j^{(T)}\}_{j=1}^{L}), \quad \hat{y}_{ij}^{\text{recon}} = \sigma(\tilde{\mathbf{z}}_i^l \cdot \mathbf{u}_j^{(T)})
\label{eq:label_recon_pred}
\end{equation}
where $\sigma(\cdot)$ is the sigmoid function. This pathway serves as a regularizer, ensuring that the aggregation in Equation~\ref{eq:label_aggregation} preserves sufficient information to reconstruct the original label assignments.

\textbf{Feature Prediction Pathway.} When processing the feature embedding $\mathbf{z}_i^f$, the decoder produces label predictions similarly:
\begin{equation}
\tilde{\mathbf{z}}_i^f = \text{CrossAttn}(\mathbf{z}_i^f; \{\mathbf{u}_j^{(T)}\}_{j=1}^{L}), \quad \hat{y}_{ij} = \sigma(\tilde{\mathbf{z}}_i^f \cdot \mathbf{u}_j^{(T)})
\label{eq:prediction}
\end{equation}
where $\hat{y}_{ij} \in [0,1]$ represents the predicted probability that label $j$ is positive for sample $i$.

\subsection{Learning Objectives}
\label{sec:learning_objectives}

HyperLabel is trained with a multi-objective loss function:
\begin{equation}
\mathcal{L}_{\text{total}} = \mathcal{L}_{\text{align}} + \lambda_1 \mathcal{L}_{\text{recon}} + \lambda_2 \mathcal{L}_{\text{sup}} + \lambda_3 \mathcal{L}_{\text{contrast}}
\label{eq:total_loss}
\end{equation}
where $\lambda_1, \lambda_2, \lambda_3$ are hyperparameters balancing the loss terms.

\textbf{Alignment Loss.} To ensure semantic consistency between feature and label representations, we minimize:
\begin{equation}
\mathcal{L}_{\text{align}} = \frac{1}{N} \sum_{i=1}^N \|\mathbf{z}_i^f - \mathbf{z}_i^l\|^2
\label{eq:align_loss}
\end{equation}
This loss encourages features and their corresponding labels to occupy nearby regions in the latent space, facilitating cross-modal information exchange.

\textbf{Label Reconstruction Loss.} To preserve label semantic information through the aggregation and decoding process, we apply binary cross-entropy between reconstructed predictions and ground-truth labels:
\begin{equation}
\mathcal{L}_{\text{recon}} = -\frac{1}{N} \sum_{i=1}^N \sum_{j=1}^L \left[y_{ij} \log \hat{y}_{ij}^{\text{recon}} + (1-y_{ij}) \log(1-\hat{y}_{ij}^{\text{recon}})\right]
\label{eq:recon_loss}
\end{equation}

\textbf{Supervised Classification Loss.} The primary classification objective employs binary cross-entropy on the predicted probabilities:
\begin{equation}
\mathcal{L}_{\text{sup}} = -\frac{1}{N} \sum_{i=1}^N \sum_{j=1}^L \left[y_{ij} \log \hat{y}_{ij} + (1-y_{ij}) \log(1-\hat{y}_{ij})\right]
\label{eq:sup_loss}
\end{equation}

\textbf{Contrastive Loss.} To enhance discriminative power in the embedding space, supervised contrastive learning~\cite{khosla2020supervised} pulls positive label embeddings closer to the decoded feature embedding while pushing negative labels away:
\begin{equation}
\mathcal{L}_{\text{contrast}} = -\frac{1}{N} \sum_{i=1}^N \frac{1}{|\mathcal{P}(\mathbf{y}_i)|} \sum_{j \in \mathcal{P}(\mathbf{y}_i)} \log \frac{\exp(\tilde{\mathbf{z}}_i^f \cdot \mathbf{u}_j^{(T)} / \tau)}{\sum_{k=1}^L \exp(\tilde{\mathbf{z}}_i^f \cdot \mathbf{u}_k^{(T)} / \tau)}
\label{eq:contrast_loss}
\end{equation}
where $\tau$ is a temperature parameter. Unlike the supervised loss that operates on predicted probabilities, the contrastive loss directly shapes the geometry of the embedding space.

\section{Experiments}
In this section, we evaluate HyperLabel through comprehensive experiments designed to address three key research questions:

\textbf{RQ1:} How does HyperLabel's multi-label classification performance compare to state-of-the-art methods?

\textbf{RQ2:} What is the impact of HyperLabel's core design components on its overall performance?

\textbf{RQ3:} How does HyperLabel provide interpretability for feature-label and label-label relationships?

\subsection{Experimental Setup}

\textbf{Datasets.} We evaluate HyperLabel on seven widely used multi-label classification benchmarks spanning diverse domains (Table~\ref{tab:datasets}), including Reuters-21578~\cite{debole2005analysis} for news categorization, Bookmarks~\cite{katakis2008multilabel} and Delicious~\cite{tsoumakas2008effective} for social tagging, Bibtex~\cite{katakis2008multilabel} for bibliographic classification, Yeast~\cite{elisseeff2001kernel} for gene function prediction, Scene~\cite{boutell2004learning} for scene classification, and SIDER~\cite{kuhn2016sider} for drug side-effect prediction. Following standard practice~\cite{bai2022gaussian}, each dataset is split into training, validation, and test sets with a ratio of 80\%, 10\%, and 10\%, respectively.

\noindent\textbf{Baselines.} We compare HyperLabel against representative baselines covering diverse modeling paradigms: the classical method ML-KNN~\cite{zhang2007ml}; embedding-based methods SLEEC~\cite{bhatia2015sparse}, C2AE~\cite{yeh2017learning}, and HARAM~\cite{benites2015haram}; the graph-based method LaMP~\cite{lanchantin2019neural}; VAE-based methods MPVAE~\cite{bai2020disentangled} and C-GMVAE~\cite{bai2022gaussian}; and the contrastive learning method MulSupCon~\cite{zhang2024multi}.

\noindent\textbf{Evaluation Metrics.} Following standard practice~\cite{zhang2013review}, we employ four complementary metrics: Example-based F1 (ebF1) evaluates sample-level prediction quality; Micro-averaged F1 (miF1) aggregates across all samples and labels, giving more weight to frequent labels; Macro-averaged F1 (maF1) averages per-label F1 scores, emphasizing performance on rare labels; Hamming Accuracy (HA) measures the fraction of correctly predicted label assignments.

\begin{table}[t]
\centering
\caption{Statistics of benchmark datasets. Cardinality denotes the average number of labels per sample, and Density represents the ratio of positive labels to total possible labels.}
\label{tab:datasets}
\begin{tabular}{lcccccl}
\toprule
\textbf{Dataset} & \textbf{\#Samples} & \textbf{\#Features} & \textbf{\#Labels} & \textbf{Card.} & \textbf{Density} & \textbf{Domain} \\
\midrule
Yeast & 2,417 & 103 & 14 & 4.24 & 0.303 & Biology \\
Scene & 2,407 & 294 & 6 & 1.07 & 0.178 & Vision \\
SIDER & 1,427 & 27 & 27 & 15.30 & 0.567 & Medical \\
Reuters & 10,789 & 23,662 & 90 & 1.23 & 0.014 & Text \\
Bibtex & 7,395 & 1,836 & 159 & 2.40 & 0.015 & Bibliography \\
Bookmarks & 87,856 & 2,150 & 208 & 2.03 & 0.010 & Social Tagging \\
Delicious & 16,105 & 500 & 983 & 19.06 & 0.019 & Social Tagging \\
\bottomrule
\end{tabular}
\end{table}

\begin{table}[t]
\centering
\caption{Performance comparison on all datasets. \textbf{Bold} indicates the best performance and \underline{underline} indicates the second best.}
\label{tab:all_metrics}

\begin{minipage}[t]{0.49\linewidth}
\centering
\textbf{(a) Example-F1}\\[0.3em]
\resizebox{\linewidth}{!}{\begin{tabular}{lccccccc}
\toprule
Method & \textit{yeast} & \textit{scene} & \textit{sider} & \textit{bkmk} & \textit{deli} & \textit{reut} & \textit{bibt} \\
\midrule
ML-KNN     & 0.618 & 0.691 & 0.738 & 0.213 & 0.259 & 0.703 & 0.182 \\
HARAM      & 0.629 & 0.717 & 0.722 & 0.216 & 0.267 & 0.711 & 0.353 \\
SLEEC      & 0.643 & 0.718 & 0.581 & 0.363 & 0.308 & 0.885 & 0.409 \\
C2AE       & 0.614 & 0.698 & 0.766 & 0.309 & 0.326 & 0.818 & 0.335 \\
LaMP       & 0.632 & 0.714 & 0.758 & 0.342 & 0.319 & \underline{0.904} & 0.414 \\
MPVAE      & 0.636 & 0.710 & \underline{0.767} & 0.369 & \textbf{0.361} & 0.894 & 0.408 \\
C-GMVAE    & 0.650 & 0.753 & 0.754 & 0.373 & 0.344 & \underline{0.904} & 0.412 \\
MulSupCon  & \underline{0.658} & \underline{0.757} & 0.757 & \underline{0.384} & \underline{0.350} & 0.861 & \underline{0.432} \\
HyperLabel & \textbf{0.665} & \textbf{0.762} & \textbf{0.772} & \textbf{0.390} & \textbf{0.361} & \textbf{0.909} & \textbf{0.433} \\
\bottomrule
\end{tabular}}
\end{minipage}
\hfill
\begin{minipage}[t]{0.49\linewidth}
\centering
\textbf{(b) Micro-F1}\\[0.3em]
\resizebox{\linewidth}{!}{\begin{tabular}{lccccccc}
\toprule
Method & \textit{yeast} & \textit{scene} & \textit{sider} & \textit{bkmk} & \textit{deli} & \textit{reut} & \textit{bibt} \\
\midrule
ML-KNN     & 0.625 & 0.667 & 0.774 & 0.181 & 0.265 & 0.680 & 0.178 \\
HARAM      & 0.635 & 0.693 & 0.756 & 0.230 & 0.274 & 0.695 & 0.365 \\
SLEEC      & 0.653 & 0.699 & 0.699 & 0.300 & 0.334 & 0.845 & 0.407 \\
C2AE       & 0.626 & 0.713 & 0.789 & 0.316 & 0.349 & 0.799 & 0.388 \\
LaMP       & 0.632 & 0.730 & 0.792 & 0.348 & 0.346 & \textbf{0.882} & \underline{0.461} \\
MPVAE      & 0.636 & 0.709 & \underline{0.798} & 0.353 & \underline{0.377} & 0.866 & 0.436 \\
C-GMVAE    & 0.658 & 0.746 & 0.795 & 0.360 & 0.360 & \underline{0.876} & 0.448 \\
MulSupCon  & \underline{0.668} & \textbf{0.757} & 0.792 & \underline{0.381} & 0.370 & 0.832 & 0.453 \\
HyperLabel & \textbf{0.671} & \underline{0.755} & \textbf{0.805} & \textbf{0.383} & \textbf{0.379} & \textbf{0.882} & \textbf{0.462} \\
\bottomrule
\end{tabular}}
\end{minipage}

\vspace{0.5em}

\begin{minipage}[t]{0.49\linewidth}
\centering
\textbf{(c) Macro-F1}\\[0.3em]
\resizebox{\linewidth}{!}{\begin{tabular}{lccccccc}
\toprule
Method & \textit{yeast} & \textit{scene} & \textit{sider} & \textit{bkmk} & \textit{deli} & \textit{reut} & \textit{bibt} \\
\midrule
ML-KNN     & 0.472 & 0.693 & \textbf{0.669} & 0.041 & 0.054 & 0.066 & 0.072 \\
HARAM      & 0.448 & 0.713 & 0.651 & 0.140 & 0.075 & 0.100 & 0.226 \\
SLEEC      & 0.425 & 0.699 & 0.594 & 0.195 & 0.143 & 0.403 & 0.293 \\
C2AE       & 0.427 & 0.728 & 0.665 & 0.232 & 0.103 & 0.363 & 0.268 \\
LaMP       & 0.408 & 0.689 & 0.617 & 0.245 & 0.181 & \underline{0.523} & 0.335 \\
MPVAE      & 0.458 & 0.725 & 0.667 & 0.282 & \underline{0.190} & 0.491 & 0.333 \\
C-GMVAE    & \underline{0.475} & \underline{0.753} & 0.667 & 0.273 & \textbf{0.193} & 0.521 & \underline{0.350} \\
MulSupCon  & 0.474 & \textbf{0.762} & 0.661 & \underline{0.284} & 0.175 & 0.428 & 0.342 \\
HyperLabel & \textbf{0.482} & \textbf{0.762} & \underline{0.668} & \textbf{0.302} & \textbf{0.193} & \textbf{0.550} & \textbf{0.370} \\
\bottomrule
\end{tabular}}
\end{minipage}
\hfill
\begin{minipage}[t]{0.49\linewidth}
\centering
\textbf{(d) Hamming Acc.}\\[0.3em]
\resizebox{\linewidth}{!}{\begin{tabular}{lccccccc}
\toprule
Method & \textit{yeast} & \textit{scene} & \textit{sider} & \textit{bkmk} & \textit{deli} & \textit{reut} & \textit{bibt} \\
\midrule
ML-KNN     & 0.784 & 0.863 & 0.717 & \textbf{0.991} & \underline{0.981} & 0.992 & 0.985 \\
HARAM      & 0.744 & 0.902 & 0.652 & \underline{0.990} & \underline{0.981} & 0.905 & \underline{0.986} \\
SLEEC      & 0.782 & 0.894 & 0.677 & 0.989 & \textbf{0.982} & \underline{0.996} & 0.981 \\
C2AE       & 0.764 & 0.893 & \underline{0.751} & \textbf{0.991} & \underline{0.981} & 0.995 & \underline{0.986} \\
LaMP       & 0.796 & 0.906 & 0.750 & \textbf{0.991} & \underline{0.981} & \underline{0.996} & \textbf{0.987} \\
MPVAE      & 0.781 & 0.898 & 0.750 & \textbf{0.991} & \underline{0.981} & \underline{0.996} & \underline{0.986} \\
C-GMVAE    & 0.789 & 0.909 & \underline{0.751} & \textbf{0.991} & \textbf{0.982} & \underline{0.996} & \textbf{0.987} \\
MulSupCon  & \textbf{0.799} & \underline{0.913} & 0.744 & \textbf{0.991} & \textbf{0.982} & 0.995 & \textbf{0.987} \\
HyperLabel & \underline{0.797} & \textbf{0.915} & \textbf{0.759} & \textbf{0.991} & \textbf{0.982} & \textbf{0.997} & \textbf{0.987} \\
\bottomrule
\end{tabular}}
\end{minipage}
\end{table}

\subsection{Multi-Label Classification Results (RQ1)}

Table~\ref{tab:all_metrics} presents comprehensive performance comparisons across all datasets and metrics. HyperLabel consistently achieves superior or competitive performance, demonstrating the effectiveness of hypergraph-based modeling for label dependencies.

\textbf{Overall Performance.} HyperLabel outperforms all baselines on example-F1 across all datasets, demonstrating strong and consistent effectiveness in modeling label dependencies across different domains: Yeast (0.665 vs. MulSupCon 0.658, +1.1\%), Scene (0.762 vs. 0.757, +0.7\%), SIDER (0.772 vs. 0.757, +2.0\%), Bookmarks (0.390 vs. 0.384, +1.6\%), Delicious (0.361 vs. 0.350, +3.1\%), Reuters (0.909 vs. 0.861, +5.6\%), and Bibtex (0.433 vs. 0.432, +0.2\%). These results validate that explicit hypergraph structure effectively captures multi-way label co-occurrence patterns across diverse application scenarios.

\textbf{Label-Based Metrics.} The advantages become more pronounced on macro-F1, which treats all labels equally and emphasizes rare label performance. HyperLabel demonstrates substantial improvements: Delicious (0.193 vs. MulSupCon 0.175, +10.3\%) and Bibtex (0.370 vs. 0.342, +8.2\%). On micro-F1, which emphasizes frequent labels, we observe consistent gains: Reuters (+6.0\%), Delicious (+2.4\%), and Bibtex (+2.0\%). These improvements suggest that structural prior knowledge from hypergraph construction helps in discovering label correlations, particularly benefiting rare labels with fewer training instances.

\textbf{Comparison with Graph and VAE Methods.} Compared to LaMP, which relies on attention without structural guidance, HyperLabel shows dramatic improvements on complex datasets: Bookmarks (0.390 vs. 0.342 ebF1, +14.0\%) and Delicious (0.361 vs. 0.319, +13.2\%), highlighting that hypergraph structure is crucial for modeling rare label correlations. Against C-GMVAE, which employs elaborate contrastive learning with careful negative sampling, HyperLabel achieves comparable or better results while relying on structural inductive bias rather than purely gradient-driven discovery. On high-cardinality label datasets like Delicious (19.06 labels/sample), our advantage is more pronounced, suggesting explicit hypergraph structure is particularly valuable when samples involve many co-occurring labels. HyperLabel also consistently outperforms MPVAE's pairwise modeling, confirming the necessity of capturing high-order dependencies.

HyperLabel achieves state-of-the-art performance across diverse datasets and evaluation metrics. These results confirm that explicit hypergraph-based modeling of sample-label correspondences provides an effective structural prior for multi-label classification, addressing limitations of methods that rely primarily on learned pairwise interactions, attention mechanisms, or contrastive objectives.

\subsection{Ablation Study (RQ2)}

To understand the contribution of HyperLabel's core components, we conduct ablation studies on three representative datasets: Bibtex (Bibliography), SIDER (Medical), and Bookmarks (Social Tagging). Table~\ref{tab:ablation} presents results for four key ablations: (1) \textbf{w/o HGNN+}, where we replace the hypergraph neural network with a simple MLP encoder for labels, eliminating structural message passing; (2) \textbf{HGNN+ $\rightarrow$ GNN}, which replaces the hypergraph neural network with a standard graph neural network; (3) \textbf{Transformer $\rightarrow$ MLP}, which replaces the Transformer-based feature encoder with an MLP; and (4) \textbf{Decoupled Decoder}, where we use separate decoders for feature and label pathways instead of parameter sharing.

\begin{table}[t]
\centering
\caption{Ablation study results on selected datasets.}
\label{tab:ablation}
\small
\begin{tabular*}{\columnwidth}{@{\extracolsep{\fill}}lccccccccc@{}}
\toprule
\multirow{2}{*}{Dataset} 
& \multicolumn{3}{c}{\textit{bibtex}} 
& \multicolumn{3}{c}{\textit{sider}} 
& \multicolumn{3}{c}{\textit{bookmarks}} \\
\cmidrule(lr){2-4} \cmidrule(lr){5-7} \cmidrule(lr){8-10}
Metric & ebF1 & miF1 & maF1 & ebF1 & miF1 & maF1 & ebF1 & miF1 & maF1 \\
\midrule
w/o HGNN+ & 0.426 & 0.449 & 0.354 & 0.767 & 0.797 & 0.665 & 0.366 & 0.360 & 0.273 \\
HGNN+ $\rightarrow$ GNN & 0.424 & 0.458 & 0.362 & 0.766 & 0.799 & 0.665 & 0.383 & 0.376 & 0.293 \\
Transformer $\rightarrow$ MLP & 0.418 & 0.444 & 0.364 & 0.765 & 0.796 & 0.665 & 0.384 & 0.377 & 0.285 \\
Decoupled Decoder & 0.424 & 0.460 & 0.350 & 0.768 & 0.801 & 0.667 & 0.374 & 0.374 & 0.278 \\
HyperLabel        & \textbf{0.433} & \textbf{0.462} & \textbf{0.370} & \textbf{0.772} & \textbf{0.805} & \textbf{0.668} & \textbf{0.390} & \textbf{0.383} & \textbf{0.302} \\
\bottomrule
\end{tabular*}
\end{table}

\textbf{Impact of Message Passing and Hypergraph Structure.} Comparing \emph{w/o HGNN+} and \emph{HGNN+ $\rightarrow$ GNN} reveals two important findings. First, removing HGNN+ causes the most severe performance degradation, indicating that message passing is crucial for learning label correlations. On Bookmarks, macro-F1 drops from 0.302 to 0.273 (-9.6\%), and example-F1 decreases from 0.390 to 0.366 (-6.2\%). On SIDER, the gain is relatively modest, possibly because SIDER has fewer samples (1,427) where structural inductive bias provides less advantage. Second, replacing HGNN+ with a GNN still leads to noticeable drops, showing that the improvement is not solely due to generic graph propagation. Instead, the hypergraph structure plays an important role in capturing high-order label relationships beyond pairwise dependencies. This advantage is particularly clear on datasets with complex label co-occurrence patterns, such as Bookmarks, where macro-F1 decreases from 0.302 to 0.293 after replacing HGNN+ with GNN. These results verify that structural message passing is effective for label correlation modeling, while hypergraph construction further strengthens the modeling of high-order dependencies.

\textbf{Impact of Shared Decoder.} Using decoupled decoders degrades performance across all metrics, though less severely than removing HGNN+. On Bookmarks, macro-F1 decreases from 0.302 to 0.278 (-7.9\%), and example-F1 drops from 0.390 to 0.374 (-4.1\%). On Bibtex, macro-F1 changes from 0.370 to 0.350 (-5.4\%), while example-F1 decreases from 0.433 to 0.424 (-2.1\%). These results validate our design rationale that parameter sharing enforces feature and label embeddings to reside in a unified semantic space, enabling effective cross-modal information exchange. Without shared parameters, the model loses the implicit information transfer mechanism where gradients from both supervised classification and reconstruction objectives jointly shape the decoder's representations.

\textbf{Impact of Transformer Feature Encoder.} Replacing the Transformer feature encoder with an MLP consistently degrades performance, indicating that the feature encoder architecture is important for learning high-quality input representations. On Bookmarks, macro-F1 drops from 0.302 to 0.285 (-5.6\%), while example-F1 decreases from 0.390 to 0.384 (-1.5\%). Similar trends on other datasets further support that the Transformer encoder better captures contextual interactions within input features, producing more informative representations for downstream multi-label prediction.

Overall, the ablation results demonstrate that all components contribute to HyperLabel's performance, with HGNN+ being particularly critical for datasets with complex label structures, highlighting the importance of structural message passing for modeling label correlations. These ablations confirm that Transformer feature encoder, hypergraph-based structural modeling and cross-modal parameter sharing are essential for effective multi-label classification.

\begin{figure}[t]
\includegraphics[width=\textwidth]{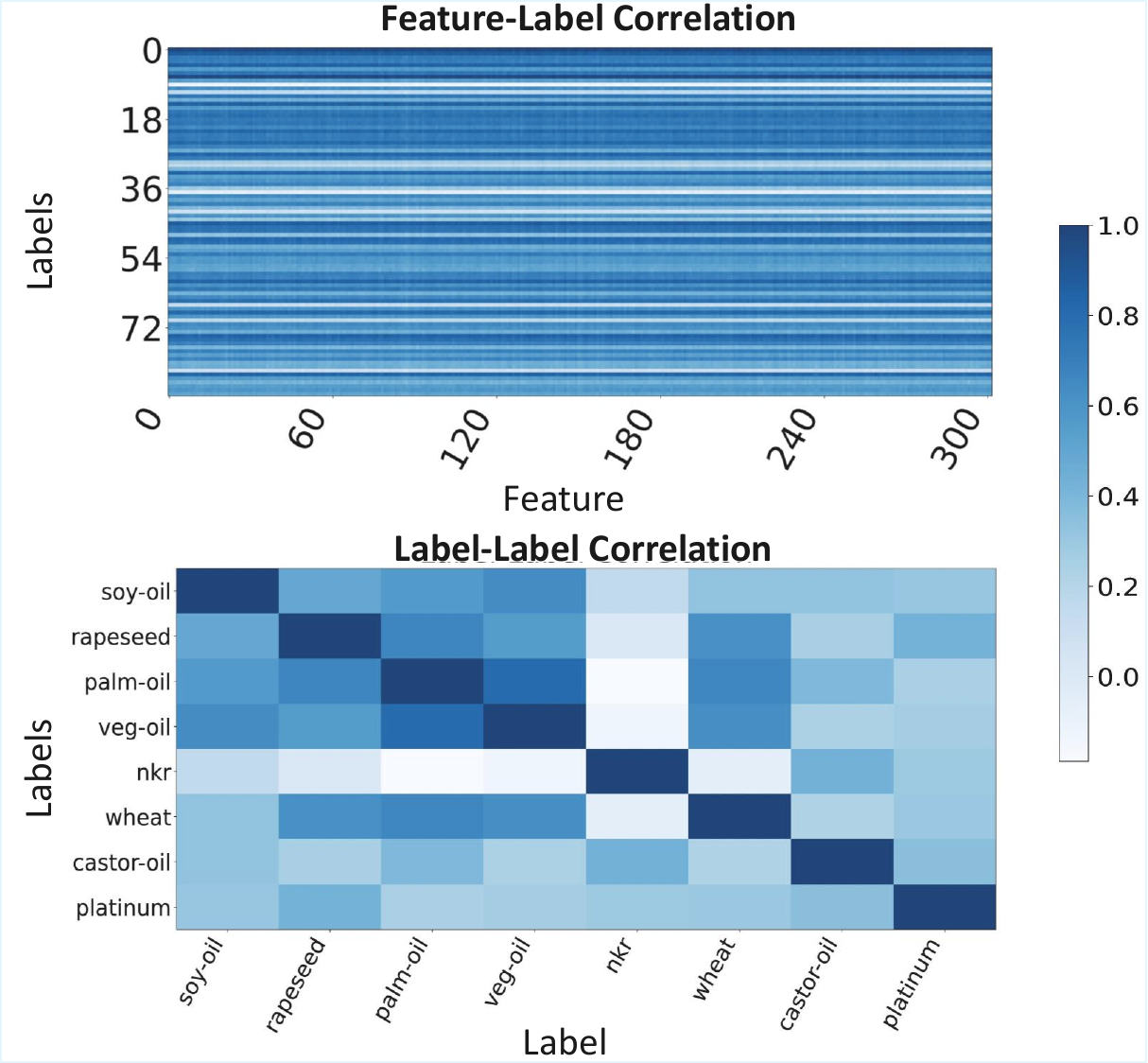}
\caption{Learned correlations in HyperLabel on a Reuters sample. \textbf{Top}: feature-label correlation map. \textbf{Bottom}: label-label correlation map.}
\label{fig:attn_map}
\end{figure}

\subsection{Interpretability Analysis (RQ3)}

HyperLabel provides interpretability through its correlation-enhanced architecture. Figure~\ref{fig:attn_map} visualizes learned correlations on a Reuters sample. The feature-label correlation map shows clear variation in how different labels respond to input features, indicating that the model captures differentiated feature-level evidence for label prediction. Such structured feature-level responses indicate that HyperLabel does not rely on uniform global signals, but instead learns differentiated associations between input semantics and candidate labels. The correlation map derived from HGNN+ message passing further reveals a coherent and interpretable structure among labels. In particular, related commodities exhibit strong co-occurrence patterns: ``soy-oil'', ``rapeseed'', ``palm-oil'', and ``veg-oil'' form a strongly correlated cluster, which is consistent with their shared semantics and frequent co-occurrence in commodity market reports. Labels such as ``wheat'' and ``castor-oil'' also show noticeable correlations with this cluster, reflecting broader agricultural and commodity-market contexts. By contrast, labels such as ``nkr'' and ``platinum'' exhibit weaker but still observable associations, likely capturing recurring co-occurrence patterns in financial news rather than direct semantic similarity. These visualizations demonstrate that HyperLabel learns interpretable representations aligned with domain knowledge, where the explicit hypergraph structure enables transparent reasoning about prediction rationale---a crucial advantage in capturing higher-order label dependencies and providing additional insight into the rationale behind multi-label predictions.

\section{Conclusion}
We presented HyperLabel, a hypergraph-based encoder-decoder framework that explicitly models complex label dependencies through structural representation. By constructing a hypergraph where sample-defined hyperedges encode multi-way label co-occurrence patterns, HyperLabel captures high-order correlations beyond pairwise interactions. The framework employs HGNN+ for bidirectional message passing to integrate feature information with label structure, and a shared cross-attention decoder with multi-objective learning to model feature-label and label-label relationships. Extensive experiments on seven benchmark datasets demonstrate state-of-the-art performance, with ablation studies confirming the critical role of hypergraph message passing and interpretability analysis revealing semantically meaningful learned correlations. Our work validates that leveraging sample-label correspondences as explicit structural inductive bias offers a principled approach to multi-label classification, particularly for tasks involving complex label semantics.

\begin{credits}
\subsubsection{\discintname}
The authors have no competing interests to declare that are relevant to the content of this article.
\end{credits}
\bibliographystyle{splncs04}
\bibliography{references}

\appendix

\section{Evaluation Metrics}

We provide detailed mathematical formulations for all evaluation metrics used in our experiments.

\textbf{Example-based F1 (ebF1).} This metric computes the F1 score for each individual sample and then averages across all samples:
\begin{equation}
\text{ebF1} = \frac{1}{N} \sum_{i=1}^{N} \frac{2 \sum_{j=1}^{L} y_{ij} \hat{y}_{ij}}{\sum_{j=1}^{L} y_{ij} + \sum_{j=1}^{L} \hat{y}_{ij}}
\end{equation}
where $y_{ij} \in \{0,1\}$ is the ground-truth label and $\hat{y}_{ij} \in \{0,1\}$ is the binary prediction after thresholding. If both the numerator and denominator are zero for a sample, we define the F1 score for that sample as 1.

\textbf{Micro-averaged F1 (miF1).} This metric aggregates true positives (TP), false positives (FP), and false negatives (FN) across all samples and labels before computing precision and recall:
\begin{align}
\text{TP} &= \sum_{i=1}^{N} \sum_{j=1}^{L} y_{ij} \hat{y}_{ij} \\
\text{FP} &= \sum_{i=1}^{N} \sum_{j=1}^{L} (1-y_{ij}) \hat{y}_{ij} \\
\text{FN} &= \sum_{i=1}^{N} \sum_{j=1}^{L} y_{ij} (1-\hat{y}_{ij})
\end{align}
Then:
\begin{align}
\text{Precision}_{\text{micro}} &= \frac{\text{TP}}{\text{TP} + \text{FP}} \\
\text{Recall}_{\text{micro}} &= \frac{\text{TP}}{\text{TP} + \text{FN}} \\
\text{miF1} &= \frac{2 \cdot \text{Precision}_{\text{micro}} \cdot \text{Recall}_{\text{micro}}}{\text{Precision}_{\text{micro}} + \text{Recall}_{\text{micro}}}
\end{align}

\textbf{Macro-averaged F1 (maF1).} This metric computes precision and recall for each label separately, then averages the F1 scores:
\begin{align}
\text{TP}_j &= \sum_{i=1}^{N} y_{ij} \hat{y}_{ij} \\
\text{FP}_j &= \sum_{i=1}^{N} (1-y_{ij}) \hat{y}_{ij} \\
\text{FN}_j &= \sum_{i=1}^{N} y_{ij} (1-\hat{y}_{ij})
\end{align}
For each label $j$:
\begin{align}
\text{Precision}_j &= \frac{\text{TP}_j}{\text{TP}_j + \text{FP}_j} \\
\text{Recall}_j &= \frac{\text{TP}_j}{\text{TP}_j + \text{FN}_j} \\
\text{F1}_j &= \frac{2 \cdot \text{Precision}_j \cdot \text{Recall}_j}{\text{Precision}_j + \text{Recall}_j}
\end{align}
Then:
\begin{equation}
\text{maF1} = \frac{1}{L} \sum_{j=1}^{L} \text{F1}_j
\end{equation}
If a label $j$ has no positive predictions or ground-truth instances, we define $\text{F1}_j = 0$.

\textbf{Hamming Accuracy (HA).} This metric measures the fraction of correctly predicted label assignments:
\begin{equation}
\text{HA} = \frac{1}{N \cdot L} \sum_{i=1}^{N} \sum_{j=1}^{L} \mathbb{I}[y_{ij} = \hat{y}_{ij}]
\end{equation}
where $\mathbb{I}[\cdot]$ is the indicator function that equals 1 when the condition is true and 0 otherwise.

\section{Computational Complexity}

\textbf{Time Complexity.} For a single forward pass, the feature encoder has complexity $\mathcal{O}(ND^2)$ where $N$ is batch size and $D$ is feature dimension. The HGNN+ encoder has complexity $\mathcal{O}(T \cdot M \cdot L \cdot d)$ where $T$ is the number of message passing iterations, $M$ is the number of unique hyperedges, $L$ is the number of labels, and $d$ is the embedding dimension. The cross-attention decoder has complexity $\mathcal{O}(N \cdot L \cdot d)$.

\textbf{Space Complexity.} The hypergraph incidence matrix $\mathbf{H} \in \mathbb{R}^{L \times M}$ and hyperedge weights $\mathbf{W} \in \mathbb{R}^{M}$ require $\mathcal{O}(L \cdot M)$ storage. Label embeddings require $\mathcal{O}(L \cdot d)$ space. The total space complexity is dominated by model parameters rather than the hypergraph structure.

\end{document}